\documentclass[10pt,twocolumn]{article}

\usepackage[margin=1in]{geometry}

\usepackage[utf8]{inputenc}
\usepackage[T1]{fontenc}
\usepackage{lmodern}
\usepackage{microtype}
\usepackage{amsmath,amssymb}
\usepackage{graphicx}
\usepackage{booktabs}
\usepackage{multirow}
\usepackage{caption}
\usepackage{subcaption}
\usepackage{xcolor}
\usepackage{enumitem}
\usepackage{tikz}
\usetikzlibrary{shapes.geometric,arrows.meta,positioning,fit,calc}
\usepackage{tabularx}
\usepackage{array}
\usepackage{float}

\usepackage[
  colorlinks=true,
  linkcolor=blue!60!black,
  citecolor=blue!60!black,
  urlcolor=blue!60!black,
  bookmarks=true,
  pdfauthor={Daniel Nobrega Medeiros},
  pdftitle={TACIT Benchmark: A Programmatic Visual Reasoning Benchmark for Generative and Discriminative Models},
]{hyperref}

\usepackage[numbers,sort&compress]{natbib}

\newcolumntype{L}[1]{>{\raggedright\arraybackslash}p{#1}}
\newcolumntype{C}[1]{>{\centering\arraybackslash}p{#1}}

\title{%
  \textbf{TACIT Benchmark: A Programmatic Visual Reasoning Benchmark\\
  for Generative and Discriminative Models}%
}

\author{%
  Daniel Nobrega Medeiros\\
  Independent Researcher\\
  \url{https://github.com/danielxmed}
}

\date{February 2026}

\begin{document}
\maketitle

\begin{abstract}
Existing visual reasoning benchmarks predominantly rely on natural language prompts, evaluate narrow reasoning modalities, or depend on subjective scoring procedures such as LLM-as-judge. We introduce the TACIT Benchmark, a programmatic visual reasoning benchmark comprising 10 tasks across 6 reasoning domains---spatial navigation, abstract pattern completion, causal simulation, logical constraint satisfaction, graph theory, and topology. The benchmark provides dual-track evaluation: a generative track in which models must produce solution images verified through deterministic computer-vision pipelines, and a discriminative track offering five-way multiple choice with structurally plausible near-miss distractors. Each distractor violates exactly one structural constraint, requiring models to reason about fine-grained visual differences rather than exploit superficial cues. Version~0.1.0 distributes 6{,}000 puzzles (108{,}000 PNG images across three resolutions) with fully deterministic seeded generation and reproducible verification. The dataset, generation code, and evaluation harness are released under the Apache~2.0 license on HuggingFace (DOI: \texttt{10.57967/hf/7904}).
\end{abstract}

\section{Introduction}
\label{sec:introduction}

The rapid advancement of multimodal foundation models~\cite{openai2023gpt4v,team2024gemini,anthropic2024claude} has created an urgent need for evaluation instruments that probe genuinely visual reasoning capabilities rather than language-mediated pattern matching. Current benchmarks fall into two broad camps: those that embed reasoning challenges in natural language with visual accompaniment~\cite{yue2024mmmu,lu2024mathvista}, and those that focus on a single reasoning modality such as abstract analogy~\cite{zhang2019raven} or general intelligence~\cite{chollet2019measure}. The former conflate linguistic competence with visual reasoning, while the latter lack the breadth needed to characterize a model's visual cognition profile across diverse domains.

A further limitation of existing benchmarks is their reliance on evaluation methods that are either narrowly discriminative (multiple choice only) or subjective (human annotators, LLM-as-judge). Discriminative-only evaluation cannot distinguish genuine solution construction from educated guessing among options, and subjective scoring introduces variability that undermines reproducibility. For visual reasoning tasks where the correct answer is a structured image---a solved maze, a colored graph, a projected silhouette---there is no principled reason why verification must involve human judgment. The answer is either structurally correct or it is not.

The TACIT Benchmark addresses these gaps through three design commitments. First, \emph{language minimality}: all task instructions are encoded visually, with no natural language clues beyond axis labels and legends, ensuring that reasoning performance reflects visual cognition rather than language comprehension. Second, \emph{dual-track evaluation}: every puzzle supports both a generative track (the model produces a solution image) and a discriminative track (the model selects from five candidates), enabling direct measurement of the gap between constructive and selective reasoning on identical stimuli. Third, \emph{deterministic verification}: every generative response is verified through a task-specific computer-vision pipeline---BFS path validation for mazes, pixel-level grid comparison for cellular automata, structural similarity for pattern matrices---eliminating evaluator subjectivity entirely.

The benchmark spans six reasoning domains instantiated in ten tasks: multi-layer maze navigation (spatial), Raven's progressive matrices (abstract pattern), forward and inverse cellular automata (causal simulation), visual logic grids (constraint satisfaction), graph $k$-coloring and isomorphism detection (graph theory), unknot detection (topology), and orthographic projection paired with isometric reconstruction (geometric projection). Each task features parameterized difficulty axes and near-miss distractors that violate exactly one structural constraint, preventing shortcut strategies based on superficial visual features.

Version~0.1.0 of the benchmark distributes 6{,}000 puzzles, producing 108{,}000 PNG images at three resolutions (512, 1024, and 2048 pixels), generated deterministically from a single seed. The contributions of this paper are:

\begin{itemize}[nosep,leftmargin=*]
  \item A benchmark of 10 visual reasoning tasks across 6 domains with parameterized difficulty.
  \item A dual-track evaluation framework with deterministic CV-based generative verification.
  \item A single-constraint distractor system ensuring structurally plausible wrong answers.
  \item An open-source generation and evaluation pipeline enabling reproducible, extensible research.
\end{itemize}

\section{Related Work}
\label{sec:related}

\paragraph{Visual reasoning benchmarks.}
Raven's Progressive Matrices~\cite{raven1938raven} have long served as a gold standard for abstract reasoning assessment. Zhang et al.~\cite{zhang2019raven} introduced RAVEN, a procedurally generated dataset of matrix reasoning problems, enabling scalable evaluation but restricted to a single reasoning modality. The Abstraction and Reasoning Corpus (ARC)~\cite{chollet2019measure,chollet2024arcprize} takes a complementary approach, emphasizing generalization from few examples across diverse grid-based tasks, though its evaluation remains discriminative-only. Bongard-LOGO~\cite{nie2020bongard} targets concept learning through visual analogy but operates on simple binary classification. These benchmarks share a common limitation: they evaluate a single facet of visual reasoning and lack generative evaluation tracks.

\paragraph{Multimodal evaluation suites.}
MMMU~\cite{yue2024mmmu} and MathVista~\cite{lu2024mathvista} evaluate multimodal models on expert-level tasks spanning dozens of disciplines. While comprehensive in scope, these benchmarks present reasoning challenges primarily through natural language, making it difficult to isolate visual reasoning from linguistic competence. CLEVR~\cite{johnson2017clevr} pioneered programmatic scene generation for visual question answering but is limited to spatial relationship queries about simple 3D scenes.

\paragraph{Programmatic and procedural benchmarks.}
Procedural generation offers several advantages for benchmark construction: unlimited scale, parameterized difficulty, deterministic reproducibility, and known ground truth. Our work builds on this tradition while extending it to encompass multiple reasoning domains, dual-track evaluation, and computer-vision-based verification of generative outputs.

\paragraph{Positioning of TACIT.}
The TACIT Benchmark distinguishes itself along four axes: (i)~\emph{language minimality}---tasks are specified visually, removing linguistic confounds; (ii)~\emph{breadth}---six reasoning domains spanning spatial, causal, logical, graph-theoretic, topological, and geometric reasoning; (iii)~\emph{dual-track evaluation}---both generative and discriminative assessment on identical puzzles; and (iv)~\emph{deterministic verification}---all scoring is performed by CV pipelines, eliminating evaluator subjectivity. Table~\ref{tab:comparison} summarizes how TACIT compares to prior work.

\begin{table}[t]
\centering
\caption{Comparison of TACIT Benchmark with related visual reasoning benchmarks. \emph{Lang-min}: language-minimal task specification. \emph{Gen}: generative evaluation track. \emph{Disc}: discriminative evaluation track. \emph{Det-V}: deterministic verification (no human/LLM judge).}
\label{tab:comparison}
\small
\begin{tabular}{@{}lcccccc@{}}
\toprule
Benchmark & Domains & Tasks & Lang-min & Gen & Disc & Det-V \\
\midrule
RAVEN & 1 & 1 & \checkmark & \texttimes & \checkmark & \checkmark \\
ARC & 1 & vary & \checkmark & \texttimes & \checkmark & \checkmark \\
Bongard-LOGO & 1 & 1 & \checkmark & \texttimes & \checkmark & \texttimes \\
MMMU & many & 900+ & \texttimes & \texttimes & \checkmark & \texttimes \\
MathVista & many & 6k+ & \texttimes & \texttimes & \checkmark & \texttimes \\
CLEVR & 1 & 1 & \texttimes & \texttimes & \checkmark & \checkmark \\
\midrule
\textbf{TACIT} & \textbf{6} & \textbf{10} & \checkmark & \checkmark & \checkmark & \checkmark \\
\bottomrule
\end{tabular}
\end{table}

\section{Benchmark Design}
\label{sec:design}

\subsection{Design Principles}

The TACIT Benchmark is organized around five design principles drawn from the theoretical framework on implicit computation in neural networks~\cite{medeiros2026tacit}:

\begin{enumerate}[nosep,leftmargin=*]
\item \textbf{Language minimality.} All task instructions are encoded through visual layout, color coding, and geometric conventions. The only text present consists of axis labels (``Layer 1'', ``State T''), numeric indices, and legend entries, ensuring that a model's performance reflects visual reasoning rather than language understanding.

\item \textbf{Dual-track evaluation.} Every puzzle instance supports both generative (produce a solution image) and discriminative (select from five candidates) evaluation, enabling researchers to quantify the gap between constructive and selective reasoning.

\item \textbf{Deterministic verification.} All generative responses are verified by task-specific computer-vision pipelines. No human judgment or LLM-as-judge is involved at any point in scoring.

\item \textbf{Parameterized difficulty.} Each task exposes independent difficulty axes (grid size, rule complexity, node count, crossing number, etc.) that control problem hardness along interpretable dimensions.

\item \textbf{Near-miss distractors.} Each distractor violates exactly one structural constraint of the correct solution, creating plausible wrong answers that cannot be eliminated by superficial feature matching.
\end{enumerate}

\subsection{Dual-Track Evaluation Architecture}

\begin{figure}[t]
\centering
\resizebox{\columnwidth}{!}{%
\begin{tikzpicture}[
    node distance=0.5cm and 0.6cm,
    box/.style={rectangle, draw, rounded corners=2pt, minimum width=1.2cm, minimum height=0.6cm, font=\scriptsize, align=center, fill=white},
    result/.style={rectangle, draw, rounded corners=2pt, minimum width=1.0cm, minimum height=0.5cm, font=\scriptsize, align=center, fill=green!10},
    arr/.style={-{Stealth[length=4pt]}, semithick},
  ]
  \node[box, fill=blue!8] (puz1) {Puzzle\\PNG};
  \node[box, right=of puz1, fill=orange!10] (model1) {Model};
  \node[box, right=of model1, fill=yellow!10] (sol1) {Solution\\PNG};
  \node[box, right=of sol1, fill=purple!10] (cv1) {CV\\Verifier};
  \node[result, right=of cv1] (res1) {Pass/\\Fail};

  \draw[arr] (puz1) -- (model1);
  \draw[arr] (model1) -- (sol1);
  \draw[arr] (sol1) -- (cv1);
  \draw[arr] (cv1) -- (res1);

  \node[above=0.1cm of model1, font=\scriptsize\bfseries] {Track 1: Generative};

  \node[box, fill=blue!8, below=1.4cm of puz1] (puz2) {Puzzle\\PNG};
  \node[box, right=0.3cm of puz2, fill=gray!10] (cands) {5 Cand.\\PNGs};
  \node[box, right=of cands, fill=orange!10] (model2) {Model};
  \node[box, right=of model2, fill=yellow!10] (sel2) {Index\\$\{0..4\}$};
  \node[result, right=of sel2] (res2) {Correct/\\Wrong};

  \draw[arr] (puz2) -- (cands);
  \draw[arr] (cands) -- (model2);
  \draw[arr] (model2) -- (sel2);
  \draw[arr] (sel2) -- (res2);

  \node[above=0.1cm of model2, font=\scriptsize\bfseries] {Track 2: Discriminative};
\end{tikzpicture}%
}
\caption{Dual-track evaluation architecture. Track~1 (generative) requires the model to produce a solution image verified by a deterministic CV pipeline. Track~2 (discriminative) presents five candidates for multiple-choice selection.}
\label{fig:evaluation-tracks}
\end{figure}
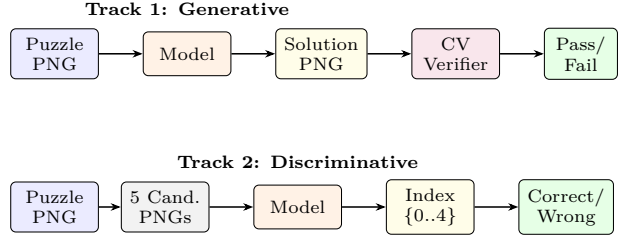

Figure~\ref{fig:evaluation-tracks} illustrates the two evaluation tracks.

\paragraph{Track~1: Generative evaluation.}
The model receives a single puzzle image and must produce a solution PNG. The solution is then passed to the task-specific verification pipeline, which performs structural validation without access to the ground-truth image (except for SSIM-based tasks). Verification produces a \texttt{VerificationResult} containing a boolean pass/fail, a human-readable reason, and a details dictionary with task-specific diagnostic information.

\paragraph{Track~2: Discriminative evaluation.}
The model receives the puzzle image together with five candidate solution images (one correct, four distractors) in randomized order. The model outputs a zero-indexed selection, and scoring is exact-match comparison. The random baseline accuracy is 20\%.

The dual-track design provides a natural diagnostic: models that perform well on Track~2 but poorly on Track~1 can discriminate correct solutions but cannot construct them---a distinction that has implications for understanding the depth of a model's visual reasoning.

\subsection{Generator Protocol and Deterministic Seeding}

Every task is implemented as a generator conforming to a shared protocol:

\begin{itemize}[nosep,leftmargin=*]
  \item \texttt{generate(difficulty, seed)} $\to$ \texttt{PuzzleInstance}: produces a complete puzzle with solution, distractors, and metadata from a deterministic seed.
  \item \texttt{verify(puzzle, candidate\_png)} $\to$ \texttt{VerificationResult}: validates a candidate PNG against the puzzle's structural constraints.
  \item \texttt{difficulty\_axes()} $\to$ \texttt{list[DifficultyRange]}: declares the task's parameterized difficulty dimensions.
\end{itemize}

All randomness flows through NumPy's \texttt{default\_rng} initialized from the puzzle seed, ensuring that every puzzle instance is fully reproducible. The release configuration uses seed~42 as the global starting seed.

\subsection{Rendering Pipeline}

Puzzles are generated as SVG drawings using a shared rendering layer that provides primitives for rectangles, circles, lines, paths, and text with consistent styling (line widths, colors, fonts). SVGs serve as the lossless source of truth and are rasterized to PNG at three resolutions---512, 1024, and 2048 pixels---via CairoSVG~\cite{cairosvg}. Track~1 verification operates entirely on PNG images, reflecting the format that models actually consume.

\subsection{Distractor System}

Each puzzle includes four distractors alongside the correct solution, yielding a five-way choice for Track~2 evaluation (20\% random baseline). The distractor generation follows a near-miss philosophy: every distractor is derived from the correct solution by introducing exactly one structural violation. Violation types are task-specific and cycle deterministically to ensure coverage across the distractor set.

This single-constraint violation design serves two purposes. First, it makes distractors structurally plausible: they resemble correct solutions and cannot be dismissed by superficial pattern matching. Second, it provides diagnostic information: if a model consistently fails on a particular violation type (e.g., wall breaches in mazes but not disconnections), this reveals specific reasoning gaps.

\section{Tasks}
\label{sec:tasks}

\begin{figure*}[t]
\centering
\begin{subfigure}[t]{0.24\textwidth}
  \includegraphics[width=\textwidth]{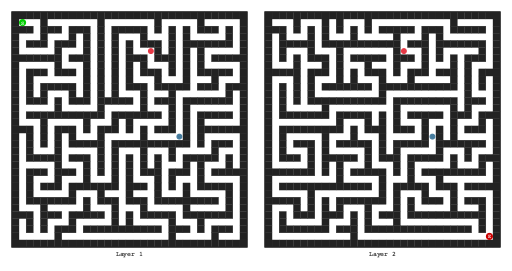}
  \caption{Maze}
\end{subfigure}\hfill
\begin{subfigure}[t]{0.24\textwidth}
  \includegraphics[width=\textwidth]{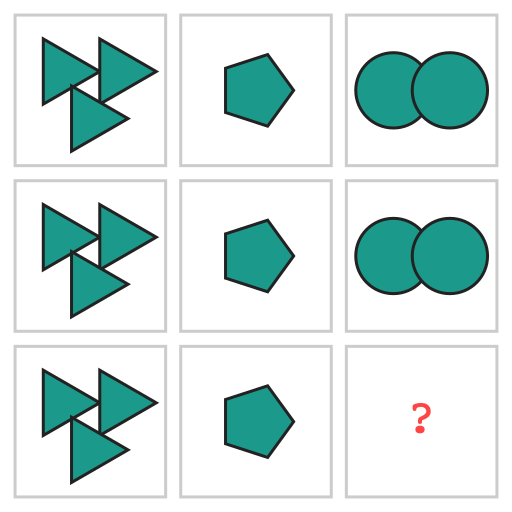}
  \caption{Raven's Matrices}
\end{subfigure}\hfill
\begin{subfigure}[t]{0.24\textwidth}
  \includegraphics[width=\textwidth]{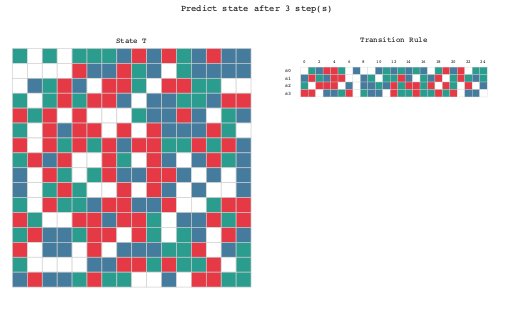}
  \caption{CA Forward}
\end{subfigure}\hfill
\begin{subfigure}[t]{0.24\textwidth}
  \includegraphics[width=\textwidth]{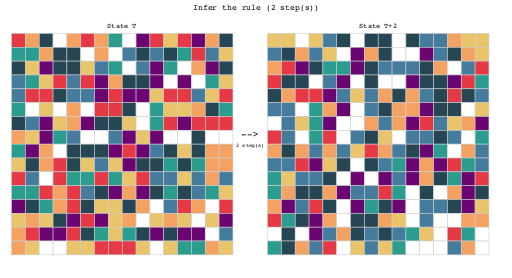}
  \caption{CA Inverse}
\end{subfigure}

\vspace{0.25cm}

\begin{subfigure}[t]{0.24\textwidth}
  \includegraphics[width=\textwidth]{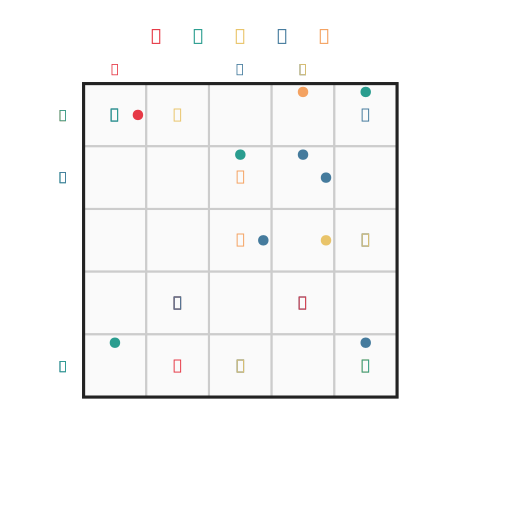}
  \caption{Logic Grid}
\end{subfigure}\hfill
\begin{subfigure}[t]{0.24\textwidth}
  \includegraphics[width=\textwidth]{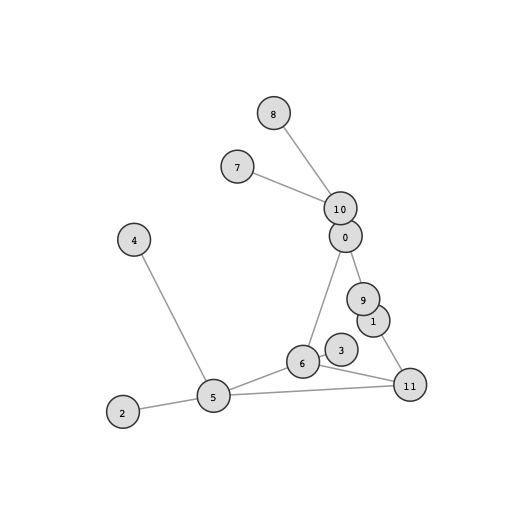}
  \caption{Graph Coloring}
\end{subfigure}\hfill
\begin{subfigure}[t]{0.24\textwidth}
  \includegraphics[width=\textwidth]{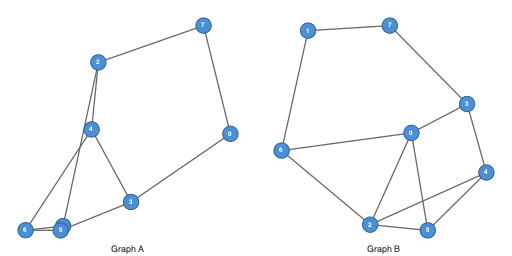}
  \caption{Graph Isomorphism}
\end{subfigure}\hfill
\begin{subfigure}[t]{0.24\textwidth}
  \includegraphics[width=\textwidth]{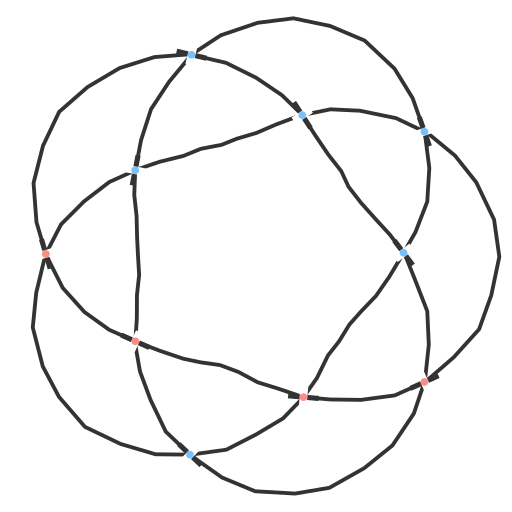}
  \caption{Unknot Detection}
\end{subfigure}

\vspace{0.25cm}

\begin{subfigure}[t]{0.24\textwidth}
  \includegraphics[width=\textwidth]{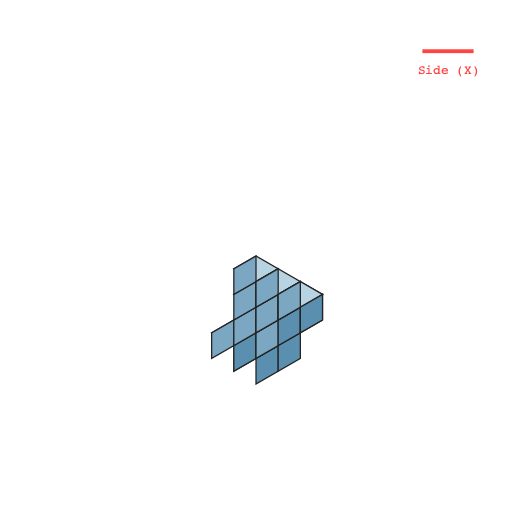}
  \caption{Ortho.\ Projection}
\end{subfigure}\hfill
\begin{subfigure}[t]{0.24\textwidth}
  \includegraphics[width=\textwidth]{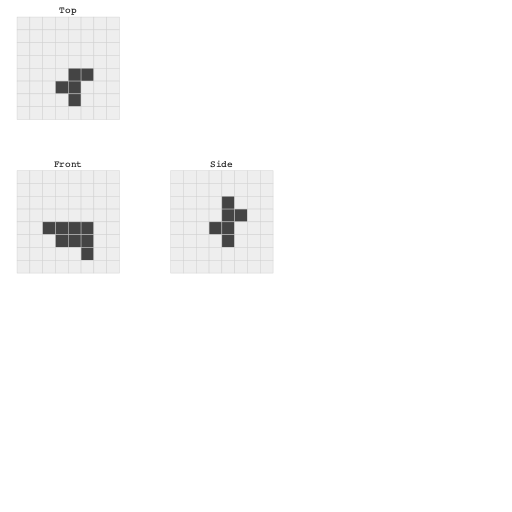}
  \caption{Iso.\ Reconstruction}
\end{subfigure}\hfill
\begin{subfigure}[t]{0.24\textwidth}
  \phantom{\includegraphics[width=\textwidth]{figures/task_01_maze_puzzle.png}}
\end{subfigure}\hfill
\begin{subfigure}[t]{0.24\textwidth}
  \phantom{\includegraphics[width=\textwidth]{figures/task_01_maze_puzzle.png}}
\end{subfigure}

\caption{Representative puzzle instances from all 10 TACIT Benchmark tasks, grouped by domain. Row~1: Spatial and Pattern tasks. Row~2: Logical, Graph, and Topology tasks. Row~3: Geometric Projection tasks. Each puzzle is specified entirely through visual encoding with minimal text annotations.}
\label{fig:hero}
\end{figure*}

The benchmark comprises 10 tasks organized into 6 reasoning domains. Table~\ref{tab:task-overview} summarizes all tasks, and Figure~\ref{fig:hero} shows representative puzzle examples.

\begin{table*}[t]
\centering
\caption{Overview of the 10 TACIT Benchmark tasks. \emph{Bin.}: binary classification task. \emph{Verification}: CV-based verification strategy. Difficulty parameters shown for easy/hard levels (200~puzzles each; medium omitted for space).}
\label{tab:task-overview}
\small
\resizebox{\textwidth}{!}{%
\begin{tabular}{@{}clllllll@{}}
\toprule
\# & Task & Domain & Bin. & Verification & Difficulty Axes & Easy & Hard \\
\midrule
1 & Multi-Layer Maze & Spatial & No & BFS path tracing & grid, layers, portals & $8{\times}8$, 1L, 0P & $32{\times}32$, 3L, 5P \\
2 & Raven's Matrices & Pattern & No & SSIM ${\geq}\,0.997$ & rules, complexity & 1 rule, additive & 3 rules, compositional \\
3 & CA Forward & Pattern & No & Pixel sampling & grid, states, steps & $8^2$, 2-state, 1-step & $32^2$, 8-state, 5-step \\
4 & CA Inverse & Pattern & No & Pixel sampling & grid, states, steps & $8^2$, 4-state, 1-step & $32^2$, 16-state, 3-step \\
5 & Logic Grid & Logical & No & Pixel sampling & grid, constraints, types & $4{\times}4$, 6C, 2T & $6{\times}6$, 16C, 4T \\
6 & Graph $k$-Coloring & Graph & No & Node sampling & nodes, density, $k$ & 6N, 0.3, $k{=}4$ & 20N, 0.5, $k{=}3$ \\
7 & Graph Isomorphism & Graph & Yes & Color counting & nodes, distortion & 5N, 0.3 & 12N, 0.9 \\
8 & Unknot Detection & Topology & Yes & Color counting & crossings & 3 & 10 \\
9 & Ortho.\ Projection & Geometric & No & Pixel sampling & faces, concavities & 6F, 0C & 16F, 3C \\
10 & Iso.\ Reconstruction & Geometric & No & SSIM ${\geq}\,0.99999$ & faces, ambiguity & 6F, 0A & 16F, 2A \\
\bottomrule
\end{tabular}%
}
\end{table*}

\subsection{Spatial Reasoning: Multi-Layer Mazes}

The maze task presents one or more 2D grid mazes arranged side-by-side, with a designated start position (green circle) on the first layer and an end position (red circle) on the last layer. In multi-layer configurations, colored portal circles at matching grid positions allow transitions between adjacent layers. The model must produce a solution image with a continuous blue path from start to end.

Verification uses a BFS-based pipeline: blue path pixels are detected, mapped to grid cells by sampling at cell center positions, and then validated through four structural checks---correct start position, correct end position, all cells lying on passable tiles, and connectivity between consecutive cells through adjacency or portal linkage. Difficulty scales along three axes: grid size (4--128), number of layers (1--8), and number of inter-layer portals (0--20).

Distractors introduce one of four violations: a path segment passing through a wall (\texttt{wall\_breach}), a cross-layer transition without a portal (\texttt{portal\_skip}), a disconnection in the path (\texttt{disconnected}), or termination at an incorrect exit (\texttt{wrong\_exit}).

\subsection{Abstract Pattern: Raven's Progressive Matrices}

Inspired by classical Raven's Progressive Matrices~\cite{raven1938raven}, this task presents a $3{\times}3$ grid of tiles with the bottom-right tile missing (marked with a red ``?''). Each tile contains geometric shapes characterized by five attributes: shape (from a pool of six), color (10-color palette), size (small/medium/large), rotation ($0^\circ$/$90^\circ$/$180^\circ$/$270^\circ$), and count (1--4 instances). Transformation rules govern how attributes change across rows and columns, with additive rules (attribute determined by column index) at lower difficulty and compositional rules (attribute determined by $(row + col) \bmod 3$) at higher difficulty.

Verification computes the Structural Similarity Index (SSIM)~\cite{wang2004image} between the candidate PNG and the ground-truth tile rendering, requiring $\text{SSIM} \geq 0.997$. This high threshold ensures that all five attributes are rendered correctly while accommodating minor rasterization differences. Each distractor perturbs exactly one attribute---shape, color, rotation, or count---creating near-miss alternatives.

\subsection{Causal Simulation: Cellular Automata}

Two complementary tasks probe causal reasoning through 2D cellular automata~\cite{wolfram2002new} with Moore neighborhoods and toroidal boundary conditions.

\paragraph{CA Forward (Task~3).}
The puzzle displays an initial grid state and a transition rule table; the model must predict the grid state after $k$ simulation steps. Verification parses the candidate PNG by sampling pixel colors at grid cell centers, mapping them to state values, and performing cell-by-cell comparison. Difficulty scales through grid size (4--64), number of cell states (2--8), and simulation steps (1--20).

\paragraph{CA Inverse (Task~4).}
The puzzle displays an initial state, a final state after $k$ steps, and the step count; the model must infer the transition rule table that maps input to output. Verification extracts the rule table from the candidate PNG and performs entry-by-entry comparison. This task is strictly harder than forward prediction, as it requires inverse reasoning: inferring the function from its input-output behavior.

\subsection{Logical Constraint Satisfaction: Visual Logic Grids}

This task presents an $N{\times}N$ grid with visual constraint clues (no natural language) using colored geometric symbols. A legend maps colors to shapes, and constraints are rendered as directional arrows (row/column placement), exclusion marks (forbidden placements), and adjacency connectors (same/different requirements). The underlying structure is a Latin square---each symbol appears exactly once per row and column~\cite{colbourn2007handbook}.

The generator validates that each puzzle's constraint set uniquely determines a solution via backtracking search. Verification extracts the grid by sampling symbol colors with inverse-distance weighted voting, then checks Latin square validity, constraint satisfaction, and exact grid match. Difficulty increases through grid size (3--8), constraint count (4--20), and number of constraint types (1--4).

\subsection{Graph Theory: Coloring and Isomorphism}

\paragraph{Graph $k$-Coloring (Task~6).}
A planar graph is presented with numbered gray nodes and edges; the model must color all nodes using exactly $k$ colors such that no two adjacent nodes share a color~\cite{diestel2017graph}. Graphs are generated via Delaunay triangulation with controlled edge pruning. Verification uses occlusion-aware node sampling---necessary because nodes may visually overlap---to extract colors and validates completeness, proper coloring, and exact $k$-color usage. Reducing $k$ toward the chromatic number at higher difficulty makes the task computationally harder.

\paragraph{Graph Isomorphism (Task~7).}
Two graphs with identical node counts but different visual layouts are displayed side-by-side. The model must determine whether they are structurally isomorphic despite different spatial arrangements. This is a binary classification task; verification counts green versus red pixels in the candidate badge image. Layout distortion (random noise added to spring-layout positions) controls difficulty by obscuring structural correspondence.

\subsection{Topology: Unknot Detection}

The unknot task presents a 2D knot diagram projection with crossing indicators (over-strand as thick line, under-strand as white gap)~\cite{adams2004knot}. Positive crossings are marked with blue dots, negative with red. The model must determine whether the diagram represents the unknot (topologically equivalent to a simple circle, simplifiable through Reidemeister moves) or a non-trivial knot. Unknots are generated by inserting Reidemeister-I loops into circular paths; non-trivial knots use torus-knot parametrizations (trefoil, figure-eight, cinquefoil, etc.). Difficulty scales with crossing count (2--15), as more crossings make visual simplification harder.

\subsection{Geometric Projection: Orthographic and Isometric}

These two tasks form a forward-inverse pair probing 3D spatial reasoning.

\paragraph{Orthographic Projection (Task~9).}
A 3D voxel solid is rendered in isometric view with an axis indicator specifying a projection direction (front, top, or side). The model must produce the corresponding 2D orthographic silhouette---a grid of filled and empty cells~\cite{bertoline2002fundamentals}. Verification samples pixel colors at grid positions and performs cell-by-cell comparison. Concavities (interior voxels removed) increase difficulty because their effect on the silhouette depends on the projection axis.

\paragraph{Isometric Reconstruction (Task~10).}
Three orthographic projections (front, top, side) are given; the model must reconstruct the 3D solid in isometric view---the inverse of Task~9. Verification uses SSIM with a threshold of $0.99999$, requiring near-exact visual reproduction. Ambiguity (redundant voxels removed) increases difficulty by creating configurations where multiple 3D shapes could produce the same three projections.

\section{Dataset Statistics and Distribution}
\label{sec:statistics}

\subsection{Scale and Composition}

The v0.1.0 release contains 6{,}000 puzzle instances: 10 tasks $\times$ 3 difficulty levels $\times$ 200 puzzles per cell. Each puzzle comprises one puzzle image, one solution image, and four distractor images, yielding 36{,}000 images per resolution. At three resolutions (512, 1024, and 2048 pixels), the total dataset contains 108{,}000 PNG files. The estimated uncompressed size is 3.9--8 GB depending on task complexity and resolution.

\subsection{Difficulty Distribution}

Every task contributes equally across difficulty levels (200 puzzles each at easy, medium, and hard), ensuring balanced evaluation. Within each difficulty level, puzzle parameters are fixed to the values specified in Table~\ref{tab:task-overview}, providing controlled comparison across tasks at matched difficulty.

\subsection{Resolution Options}

Three PNG resolutions serve different use cases: 512px for rapid prototyping and models with limited visual input resolution, 1024px as the recommended default for standard evaluation, and 2048px for high-fidelity analysis and models that benefit from fine-grained visual detail. All three resolutions are deterministically rasterized from the same SVG source, ensuring pixel-perfect correspondence of content.

\subsection{Distribution and Reproducibility}

The dataset is hosted on HuggingFace as \texttt{tylerxdurden/TACIT-benchmark} with DOI \texttt{10.57967/hf/7904}~\cite{medeiros2026tacitbenchmark}. The hierarchical directory structure organizes images by task, difficulty, and resolution:

\begin{center}
\small
\texttt{task\_\{NN\}\_\{name\}/\{difficulty\}/\{resolution\}/}
\end{center}

Each puzzle includes a JSON metadata file containing the task name, seed, difficulty parameters, distractor violation types, and all information needed to regenerate the puzzle from source. Combined with the deterministic generation pipeline (global seed~42, NumPy \texttt{default\_rng}), the entire dataset is reproducible from the release configuration alone. The generation code and evaluation harness are released under the Apache~2.0 license.

\section{Evaluation Protocol}
\label{sec:evaluation}

\subsection{Track~1: Generative Evaluation}

For generative evaluation, the model receives puzzle PNG files and must produce corresponding solution PNGs following the naming convention \texttt{\{task\}\_\{difficulty\}\_\{seed\}.png}. The evaluation harness loads each puzzle's metadata, regenerates the generator instance, and calls \texttt{verify(puzzle, candidate\_png)} to obtain a \texttt{VerificationResult}.

\begin{figure}[t]
\centering
\begin{tikzpicture}[
    node distance=0.6cm,
    box/.style={rectangle, draw, rounded corners=2pt, minimum width=2.2cm, minimum height=0.55cm, font=\scriptsize, align=center, fill=white},
    arr/.style={-{Stealth[length=4pt]}, thick},
  ]
  \node[box, fill=blue!8] (gen) {SVG\\Generation};
  \node[box, right=of gen, fill=green!8] (rast) {CairoSVG\\Rasterization};
  \node[box, right=of rast, fill=orange!8] (png) {Multi-Res\\PNG Output};

  \draw[arr] (gen) -- (rast);
  \draw[arr] (rast) -- (png);

  \node[box, below=1.0cm of png, fill=purple!8] (cv) {CV\\Verification};
  \node[box, left=of cv, fill=yellow!8] (cand) {Candidate\\PNG};

  \draw[arr] (png) -- node[right, font=\scriptsize]{ground truth} (cv);
  \draw[arr] (cand) -- (cv);

  \node[above=0.1cm of rast, font=\scriptsize\bfseries] {Generation Pipeline};
\end{tikzpicture}
\caption{Generation and verification pipeline. SVGs are generated deterministically, rasterized to multi-resolution PNGs, and verified against candidates using task-specific CV pipelines.}
\label{fig:pipeline}
\end{figure}
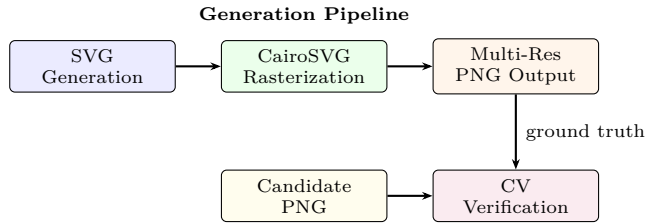

The verification methods are task-specific and fall into three categories (Figure~\ref{fig:pipeline}):

\begin{enumerate}[nosep,leftmargin=*]
  \item \textbf{Structural pixel sampling} (Tasks 1, 3--6, 9): Grid cells or node positions are sampled from the candidate PNG, colors are mapped to structural values (states, symbols, fill colors), and the extracted structure is validated against task constraints.
  \item \textbf{Color counting} (Tasks 7--8): Green and red pixel counts determine binary classification answers for graph isomorphism and unknot detection.
  \item \textbf{SSIM comparison} (Tasks 2, 10): Structural similarity between candidate and ground-truth PNGs, with task-specific thresholds ($\geq 0.997$ for Raven, $\geq 0.99999$ for isometric reconstruction).
\end{enumerate}

\subsection{Track~2: Discriminative Evaluation}

For discriminative evaluation, the model receives a puzzle PNG and five candidate solution PNGs (one correct solution and four distractors in randomized order). The model must output the zero-indexed position of the correct solution. Results are submitted as a JSON file with fields for puzzle ID, task name, difficulty, correct index, and selected index. Scoring is exact-match accuracy.

\subsection{Scoring and Aggregation}

Performance is reported at three levels of granularity:

\begin{itemize}[nosep,leftmargin=*]
  \item \textbf{Per-task accuracy}: fraction of correct responses within each task.
  \item \textbf{Per-domain accuracy}: average accuracy across tasks within each reasoning domain.
  \item \textbf{Cross-track gap}: difference between Track~2 (discriminative) and Track~1 (generative) accuracy for each task, quantifying the constructive-selective reasoning gap.
\end{itemize}

Results can be further disaggregated by difficulty level to produce difficulty-stratified profiles revealing how model performance degrades with increasing task complexity.

\section{Puzzle--Solution Examples}
\label{sec:examples}

\begin{figure*}[t]
\centering
\begin{subfigure}[t]{0.24\textwidth}
  \includegraphics[width=\textwidth]{figures/task_01_maze_puzzle.png}
  \caption{Maze puzzle}
\end{subfigure}\hfill
\begin{subfigure}[t]{0.24\textwidth}
  \includegraphics[width=\textwidth]{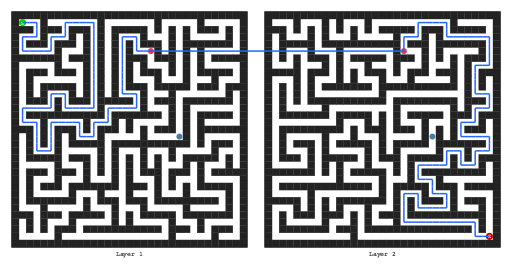}
  \caption{Maze solution}
\end{subfigure}\hfill
\begin{subfigure}[t]{0.24\textwidth}
  \includegraphics[width=\textwidth]{figures/task_06_graph_coloring_puzzle.png}
  \caption{Graph coloring puzzle}
\end{subfigure}\hfill
\begin{subfigure}[t]{0.24\textwidth}
  \includegraphics[width=\textwidth]{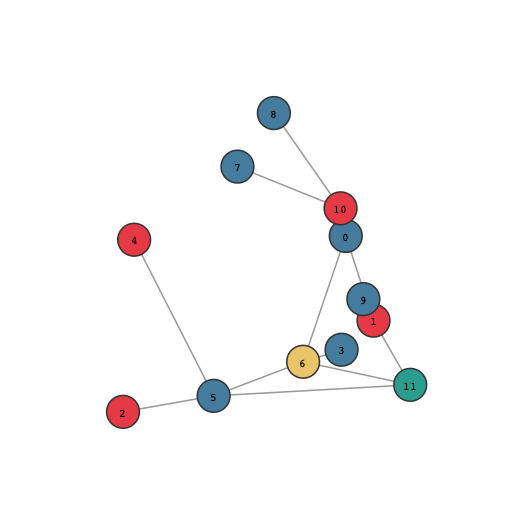}
  \caption{Graph coloring solution}
\end{subfigure}

\vspace{0.3cm}

\begin{subfigure}[t]{0.24\textwidth}
  \includegraphics[width=\textwidth]{figures/task_09_ortho_projection_puzzle.png}
  \caption{Ortho.\ projection puzzle}
\end{subfigure}\hfill
\begin{subfigure}[t]{0.24\textwidth}
  \includegraphics[width=\textwidth]{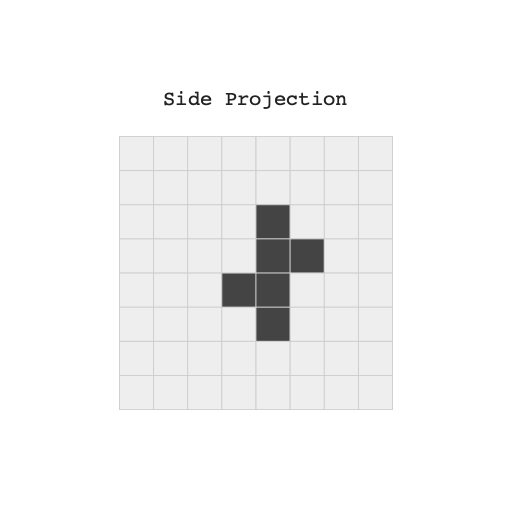}
  \caption{Ortho.\ projection solution}
\end{subfigure}\hfill
\begin{subfigure}[t]{0.24\textwidth}
  \includegraphics[width=\textwidth]{figures/task_02_raven_puzzle.png}
  \caption{Raven puzzle}
\end{subfigure}\hfill
\begin{subfigure}[t]{0.24\textwidth}
  \includegraphics[width=\textwidth]{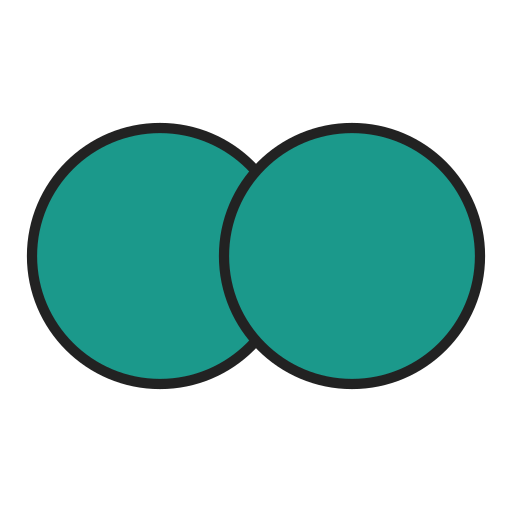}
  \caption{Raven solution}
\end{subfigure}

\caption{Puzzle--solution pairs for four representative tasks. \textbf{(a--b)}~Multi-layer maze with blue path from start (green) to end (red). \textbf{(c--d)}~Graph $k$-coloring: gray uncolored nodes and their proper coloring. \textbf{(e--f)}~Orthographic projection: 3D voxel solid and its 2D silhouette. \textbf{(g--h)}~Raven's matrix with missing tile completed.}
\label{fig:puzzle-solution}
\end{figure*}

Figure~\ref{fig:puzzle-solution} presents puzzle--solution pairs for four representative tasks, illustrating the visual format of both inputs and expected outputs across different reasoning domains.

\section{Limitations and Future Work}
\label{sec:limitations}

This release (v0.1.0) is the initial dataset announcement; several limitations will be addressed in future versions.

\paragraph{No baseline results.}
We do not report model performance in this paper. While the evaluation harness is fully functional, comprehensive baseline experiments across multiple multimodal models require significant compute resources and are planned for a follow-up study. We encourage the research community to benchmark their models using the provided infrastructure.

\paragraph{No human performance calibration.}
The difficulty levels (easy, medium, hard) are defined by parameter ranges chosen based on the authors' judgment and preliminary qualitative assessment, not empirical calibration against human performance. A formal human baseline study would establish the relationship between parameterized difficulty and actual task hardness.

\paragraph{Fixed distractor count.}
All puzzles include exactly four distractors (five-way choice). Future versions may explore variable distractor counts to study how discrimination performance scales with the number of alternatives.

\paragraph{Limited task scope.}
While six reasoning domains provide broader coverage than prior benchmarks, important reasoning modalities---temporal reasoning, physical simulation, counting under occlusion, compositional scene understanding---remain uncovered. The modular generator architecture facilitates adding new tasks without modifying the evaluation infrastructure.

\paragraph{Documentation language.}
Documentation is currently available in English and Chinese. Expanding to additional languages would increase global accessibility.

\paragraph{Future directions.}
Planned extensions include: (i)~baseline evaluation of frontier multimodal models, (ii)~human performance calibration, (iii)~difficulty auto-calibration using psychometric item response theory, (iv)~additional tasks and reasoning domains, and (v)~a public leaderboard for community benchmarking.

\section{Broader Impact}
\label{sec:impact}

The TACIT Benchmark is designed to advance the scientific understanding of visual reasoning capabilities in multimodal AI systems. By providing language-minimal tasks with deterministic verification, it enables evaluation that is both more targeted (isolating visual reasoning from linguistic ability) and more reproducible (eliminating evaluator subjectivity) than existing alternatives.

The dual-track evaluation design provides a diagnostic tool for understanding the depth of model reasoning: the gap between generative and discriminative performance on identical stimuli offers a quantitative measure of whether a model can construct solutions or merely recognize them. This distinction has practical relevance for applications ranging from automated design to scientific visualization.

We note that the benchmark evaluates reasoning about abstract visual structures and does not involve human subjects, personal data, or safety-critical decision-making. The programmatic generation approach ensures that all puzzle content is synthetic and controlled, posing no privacy or fairness concerns. The open-source release under Apache~2.0 enables broad access and community extension.

\section{Conclusion}

We have presented the TACIT Benchmark, a programmatic visual reasoning benchmark comprising 10 tasks across 6 reasoning domains with dual-track evaluation and deterministic computer-vision verification. The benchmark addresses fundamental limitations of existing evaluation instruments: it eliminates linguistic confounds through language-minimal task design, enables measurement of constructive reasoning through generative evaluation, and ensures reproducible scoring through fully automated verification pipelines. The v0.1.0 release distributes 6{,}000 puzzles (108{,}000 images) on HuggingFace with complete generation and evaluation code, providing an immediate resource for researchers evaluating multimodal models. We invite the community to benchmark their models using TACIT and to extend the benchmark with additional tasks and reasoning domains.

\section*{Acknowledgments}

The TACIT Benchmark builds on the theoretical framework introduced in \cite{medeiros2026tacit}. We thank the open-source communities behind CairoSVG, OpenCV~\cite{bradski2000opencv}, scikit-image~\cite{vanderwalt2014scikit}, and HuggingFace Datasets~\cite{lhoest2021datasets} for the tools that made this work possible.

\bibliographystyle{plainnat}
\bibliography{references}

\appendix

\section{Verification Details}
\label{app:verification}

Table~\ref{tab:verification} provides detailed verification specifications for each task.

\begin{table}[h!]
\centering
\caption{Verification specifications for all 10 tasks.}
\label{tab:verification}
\scriptsize
\setlength{\tabcolsep}{3pt}
\begin{tabular}{@{}clll@{}}
\toprule
\# & Strategy & Threshold & Failure Diagnostic \\
\midrule
1 & BFS path tracing & 4 structural checks & Failed check ID \\
2 & SSIM & $\geq 0.997$ & SSIM score \\
3 & Pixel sampling & All cells match & Diff.\ cells / total \\
4 & Pixel sampling & All entries match & Diff.\ entries / total \\
5 & Color voting & Latin sq.\ + exact & Constraint ID \\
6 & Node sampling & Proper $k$-coloring & Violating node(s) \\
7 & Color counting & Majority matches & Answer vs.\ expected \\
8 & Color counting & Majority matches & Answer vs.\ expected \\
9 & Pixel sampling & All cells match & Diff.\ cells / total \\
10 & SSIM & $\geq 0.99999$ & SSIM score \\
\bottomrule
\end{tabular}
\end{table}

\section{Distractor Violation Types}
\label{app:distractors}

Each distractor violates exactly one structural constraint. Table~\ref{tab:distractors} lists violation types by task.

\begin{table}[h!]
\centering
\caption{Distractor violation types per task.}
\label{tab:distractors}
\scriptsize
\setlength{\tabcolsep}{3pt}
\begin{tabular}{@{}cL{6.5cm}@{}}
\toprule
\# & Violation Types \\
\midrule
1 & \texttt{wall\_breach}, \texttt{portal\_skip}, \texttt{disconnected}, \texttt{wrong\_exit} \\
2 & \texttt{wrong\_shape}, \texttt{wrong\_color}, \texttt{wrong\_rotation}, \texttt{wrong\_count} \\
3 & \texttt{wrong\_cell}, \texttt{wrong\_step\_count}, \texttt{wrong\_rule} \\
4 & \texttt{off\_by\_one\_rule}, \texttt{transposed\_rule}, \texttt{partial\_rule} \\
5 & \texttt{constraint\_violation}, \texttt{symbol\_swap}, \texttt{non\_unique} \\
6 & \texttt{adjacent\_conflict}, \texttt{missing\_color}, \texttt{wrong\_k} \\
7 & \texttt{opposite\_answer} \\
8 & \texttt{opposite\_answer} \\
9 & \texttt{wrong\_axis}, \texttt{missing\_feature}, \texttt{extra\_feature}, \texttt{mirrored} \\
10 & \texttt{wrong\_depth}, \texttt{missing\_face}, \texttt{extra\_volume}, \texttt{rotated} \\
\bottomrule
\end{tabular}
\end{table}

\section{Release Configuration}
\label{app:config}

The v0.1.0 release is generated from \texttt{configs/release.yaml} with global seed~42. Table~\ref{tab:release-config} shows the difficulty parameters.

\begin{table}[h!]
\centering
\caption{Release parameters (200~puzzles per task per level).}
\label{tab:release-config}
\scriptsize
\setlength{\tabcolsep}{3pt}
\begin{tabular}{@{}cllll@{}}
\toprule
\# & Task & Easy & Medium & Hard \\
\midrule
1 & Maze & $8^2$, 1L, 0P & $16^2$, 2L, 2P & $32^2$, 3L, 5P \\
2 & Raven & 1r, add. & 2r, add. & 3r, comp. \\
3 & CA Fwd & $8^2$, 2s, 1t & $16^2$, 4s, 3t & $32^2$, 8s, 5t \\
4 & CA Inv & $8^2$, 4s, 1t & $16^2$, 8s, 2t & $32^2$, 16s, 3t \\
5 & Logic & $4^2$, 6c, 2ty & $5^2$, 10c, 3ty & $6^2$, 16c, 4ty \\
6 & Color & 6n, .3, $k{=}4$ & 12n, .4, $k{=}4$ & 20n, .5, $k{=}3$ \\
7 & Iso Det & 5n, .3$\delta$ & 8n, .6$\delta$ & 12n, .9$\delta$ \\
8 & Unknot & 3 cr & 6 cr & 10 cr \\
9 & Ortho & 6f, 0c & 10f, 1c & 16f, 3c \\
10 & Iso Rec & 6f, 0a & 10f, 1a & 16f, 2a \\
\bottomrule
\end{tabular}
\end{table}

\end{document}